\documentclass[11pt,twoside]{article} 

\usepackage[utf8]{inputenc}
\usepackage{graphicx}
\usepackage{amsmath, amssymb, bm}
\usepackage{hyperref}
\usepackage{natbib}
\usepackage{fancyhdr}
\usepackage{geometry}
\usepackage{titlesec}
\usepackage{lipsum}
\usepackage{orcidlink} %
\usepackage{booktabs}
\usepackage{placeins}
\usepackage{float} 

\setcitestyle{authoryear}
\hypersetup{
	colorlinks=true,
	linkcolor=blue,
	filecolor=blue,
	citecolor=blue,
	urlcolor=black
}

\fancypagestyle{plain}{  
	\fancyhf{}
	
}

\title{
	\vspace{1em}
	\hrule height 1.5pt
	\vspace{0.4em}
	Half-AVAE: Adversarial-Enhanced Factorized and Structured Encoder-Free VAE for Underdetermined Independent Component Analysis
	\vspace{0.8em}
	\hrule height 1.5pt
	\vspace{1em}
}

\author{
	\begin{minipage}{.3\textwidth}
		\centering
		\small
		\textbf{Yuan-Hao Wei\orcidlink{0000-0001-9439-0780}}\\
		Hong Kong Polytechnic University \\
		\texttt{Yuan-Hao.Wei@outlook.com}
	\end{minipage}
	\hfill
	\begin{minipage}{.3\textwidth}
		\centering
		\small
		\textbf{Yan-Jie Sun\orcidlink{0000-0002-7967-6382}}\\
		Hong Kong Polytechnic University \\
		\texttt{Yanjie.Sun@connect.polyu.hk}
	\end{minipage}
}
\date{}

\begin{document}
	\maketitle
	\thispagestyle{plain} 
	
	\begin{abstract}
    This study advances the Variational Autoencoder (VAE) framework by addressing challenges in Independent Component Analysis (ICA) under both determined and underdetermined conditions, focusing on enhancing the independence and interpretability of latent variables. Traditional VAEs map observed data to latent variables and back via an encoder-decoder architecture, but struggle with underdetermined ICA where the number of latent variables exceeds observed signals. The proposed Half Adversarial VAE (Half-AVAE) builds on the encoder-free Half-VAE framework, eliminating explicit inverse mapping to tackle underdetermined scenarios. By integrating adversarial networks and External Enhancement (EE) terms, Half-AVAE promotes mutual independence among latent dimensions, achieving factorized and interpretable representations. Experiments with synthetic signals demonstrate that Half-AVAE outperforms baseline models, including GP-AVAE and Half-VAE, in recovering independent components under underdetermined conditions, as evidenced by lower root mean square errors. The study highlights the flexibility of VAEs in variational inference, showing that encoder omission, combined with adversarial training and structured priors, enables effective solutions for complex ICA tasks, advancing applications in disentanglement, causal inference, and generative modeling.
	\end{abstract}
	
	\section{Introduction}
    The Variational Autoencoder (VAE) (\cite{kingma2013auto}; \cite{kingma2019introduction}) integrates the variational Bayesian theory (\cite{rezende2014stochastic}) with an encoder-decoder architecture (\cite{rumelhart1986learning}; \cite{bourlard1988auto}; \cite{rumelhart1986learning}). The encoding process can be conceptualized as a mapping from observed data $\mathbf{X}$ to latent variables $\mathbf{Z}$, that is, $\mathbf{Z} = f^{-1}(\mathbf{X})$ (\cite{wei2024innovative}; \cite{wei2024half}), or equivalently as the calculation of posterior probability $P(\mathbf{Z}|\mathbf{X})$ (\cite{wei2024innovative}). In contrast, the decoding process maps latent variables $\mathbf{Z}$ back to observed data $\mathbf{X}$, that is, $\mathbf{X} = f(\mathbf{Z})$, or computes conditional probability $P(\mathbf{X}|\mathbf{Z})$ (get the posterior probability of $\mathbf{Z}$ given $\mathbf{X}$. This forms a closed-loop system that transforms observed data $\mathbf{X}$ into latent variables $\mathbf{Z}$ and back to $\mathbf{X}$.

    The interpretation of latent variables $\mathbf{Z}$ fundamentally shapes the application of VAEs. When $\mathbf{Z}$ is treated as a low-dimensional representation of $\mathbf{X}$, VAEs facilitate tasks such as feature extraction or clustering (\cite{way2018extracting}; \cite{lopez2018deep}; \cite{wang2018vasc}; \cite{eraslan2019single}; \cite{gronbech2020scvae}). However, when emphasis is placed on the disentanglement/interpretability (\cite{higgins2017beta}; \cite{chen2018isolating}; \cite{burgess2018understanding}; \cite{yang2021causalvae}; \cite{lachapelle2022disentanglement}) of the model and causal inference (\cite{pearl2019seven}; \cite{lippe2022citris}; \cite{scholkopf2022causality}; \cite{ahuja2023interventional}), $\mathbf{Z}$ can be regarded as latent factors that influence $\mathbf{X}$. Disentangling these factors into independent components $\mathbf{Z} = \{\mathbf{Z}^a, \mathbf{Z}^b, \dots, \mathbf{Z}^n\}$ improves the interpretability (\cite{chen2016infogan}; \cite{higgins2017beta};   \cite{wei2024half}), controllability (\cite{yang2021causalvae}; \cite{komanduri2022scm}), scalability (\cite{hsu2018scalable}), and generative capabilities (\cite{wei2025vaes}) of the model.
    
    Independent Component Analysis (ICA) (\cite{hyvarinen2019nonlinear}; \cite{khemakhem2020variational}), a more specific yet direct problem within this research domain, serves as a preliminary validation method. If observed data $\mathbf{X}$ are considered a mixture of independent latent factors $\mathbf{Z}$, it can be generally understood as being generated (actually mixed) by $\mathbf{Z}$. The characteristics of $\mathbf{X}$ are governed by independent components $\mathbf{Z} = \{\mathbf{Z}^a, \mathbf{Z}^b, \dots, \mathbf{Z}^n\}$ . Thus, ICA can be subsumed in the general study of interpretable generative models.
    
    Typically, the dimensionality ${n}$ of the latent space $\mathbf{Z}$ is significantly smaller than that of the observed data $\mathbf{X}$ (e.g., images, videos, or audio), denoted as $m$, i.e., $n << m$. In this case, the encoding process, represented by the inverse mapping $\mathbf{Z} = f^{-1}(\mathbf{X})$, is overdetermined, resulting in information compression. This overdetermined nature implies that the inverse mapping is information-complete, and solving such problems in ICA is generally straightforward. However, when $n > m$, the encoding process becomes underdetermined, rendering the explicit inverse mapping $\mathbf{Z} = f^{-1}(\mathbf{X})$ infeasible (\cite{comon1994independent}; \cite{cardoso1998blind}; \cite{hyvarinen2001independent}). To address this, the Half-VAE framework abandons the encoder, bypassing explicit inverse mapping and providing theoretical feasibility to address underdetermined ICA problems (\cite{wei2024half}).
    
    However, simply removing the explicit inverse mapping does not fully resolve the underdetermined ICA challenges (\cite{wei2024half}). To address this, additional mechanisms are introduced in Half-VAE to encourage independence among the dimensions of $\mathbf{Z} = \{\mathbf{Z}^a, \mathbf{Z}^b, \dots, \mathbf{Z}^n\}$. \cite{brakel2017learning} proposed the use of an adversarial network to promote independence among latent dimensions by aligning the joint distribution of signals with their factorized marginal distributions. Building on this, we combined adversarial training with GP-VAE, introducing GP-Adversarial VAE (GP-AVAE), which improves the separation of latent dimensions into independent components (\cite{wei2024innovative}).  Similarly, by integrating adversarial networks with Half-VAE, we propose the Half Adversarial VAE (Half-AVAE).
    
    To evaluate performance, we designed experiments with synthetic source signals that exhibited varied temporal structures, spanning determined and underdetermined conditions. We compared GP-AVAE, Half-VAE, and Half-AVAE. The results indicate that Half-AVAE excels in separating signals under underdetermined conditions, closely approximating source signals. Half-VAE performs worse, while GP-AVAE, which retains an explicit encoder for inverse mapping $f^{-1}$, produces the least favorable outcomes.

    \section{Methodology}

    The architecture of a Variational Autoencoder (VAE) comprises an encoder and a decoder. The encoder maps the input $\mathbf{X}$ to latent variables $\mathbf{Z}$, while the decoder reconstructs the input, denoted as $\hat{\mathbf{X}}$, from $\mathbf{Z}$. If the dimensions of the latent variables $\mathbf{Z}$, denoted as $\{\mathbf{Z}^a, \mathbf{Z}^b, \dots, \mathbf{Z}^n\}$, are treated as independent components, the encoding process can be likened to separating mixed signals $\mathbf{X}$ into independent components $\mathbf{Z}$, represented as the inverse mapping $\mathbf{Z} = f^{-1}(\mathbf{X})$. Conversely, the decoding process corresponds to recombining these independent components $\mathbf{Z}$ into the reconstructed mixed signal $\hat{\mathbf{X}}$.
    
    From a probabilistic perspective, computing the posterior probability $P(\mathbf{Z}|\mathbf{X})$ is analogous to the inverse process in Independent Component Analysis (ICA), expressed as $\mathbf{Z} = f^{-1}(\mathbf{X})$ (\cite{wei2024innovative}). Specifically, $P(\mathbf{Z}|\mathbf{X})$ represents the probability of $\mathbf{Z}$ given the observed $\mathbf{X}$, which parallels the task in ICA or general inverse problems of recovering independent components $\mathbf{Z}$ from a known mixed signal $\mathbf{X}$. However, directly obtaining the posterior $P(\mathbf{Z}|\mathbf{X})$ via Bayes' theorem is often intractable. Consequently, variational inference provides a practical approach by introducing a flexible distribution $q(\mathbf{Z}|\mathbf{X})$ to approximate the true posterior $P(\mathbf{Z}|\mathbf{X})$.

    \subsection{Derivation of VAE}
    
    The log marginal likelihood of the data is given by:
    \begin{equation}
    \ln P(\mathbf{X}) = \ln \int P(\mathbf{X}, \mathbf{Z}) \, d\mathbf{Z}. \tag{1} 
    \end{equation}
    
    Since \(\int q(\mathbf{Z}|\mathbf{X}) \, d\mathbf{Z} = 1\), the log marginal likelihood can be rewritten to incorporate \(q(\mathbf{Z}|\mathbf{X})\):
    \begin{equation}
    \ln P(\mathbf{X}) = \ln \left[ P(\mathbf{X}) \int q(\mathbf{Z}|\mathbf{X}) \, d\mathbf{Z} \right] = \int q(\mathbf{Z}|\mathbf{X}) \ln P(\mathbf{X}) \, d\mathbf{Z}. \tag{2} 
    \end{equation}
    
    Using Bayes' theorem, \(P(\mathbf{X})\) can be expressed as:
    \begin{equation}
    P(\mathbf{X}) = \frac{P(\mathbf{X}, \mathbf{Z})}{P(\mathbf{Z}|\mathbf{X})}. \tag{3} 
    \end{equation}
    
    Substituting (3) into (2), we obtain:
    \begin{align}
    \ln P(\mathbf{X}) &= \int q(\mathbf{Z}|\mathbf{X}) \ln \left( \frac{P(\mathbf{X}, \mathbf{Z})}{P(\mathbf{Z}|\mathbf{X})} \right) d\mathbf{Z} \notag \\
    &= \int q(\mathbf{Z}|\mathbf{X}) \ln \left( \frac{P(\mathbf{X}, \mathbf{Z}) q(\mathbf{Z}|\mathbf{X})}{P(\mathbf{Z}|\mathbf{X}) q(\mathbf{Z}|\mathbf{X})} \right) d\mathbf{Z} \notag \\
    &= \int q(\mathbf{Z}|\mathbf{X}) \ln \left( \frac{P(\mathbf{X}, \mathbf{Z})}{q(\mathbf{Z}|\mathbf{X})} \right) d\mathbf{Z} + \int q(\mathbf{Z}|\mathbf{X}) \ln \left( \frac{q(\mathbf{Z}|\mathbf{X})}{P(\mathbf{Z}|\mathbf{X})} \right) d\mathbf{Z} \notag \\
    &= \int q(\mathbf{Z}|\mathbf{X}) \ln \left( \frac{P(\mathbf{X}, \mathbf{Z})}{q(\mathbf{Z}|\mathbf{X})} \right) d\mathbf{Z} + \text{KL}(q(\mathbf{Z}|\mathbf{X}) \| P(\mathbf{Z}|\mathbf{X})). \tag{4} 
    \end{align}
    
    The objective is to make \(q(\mathbf{Z}|\mathbf{X})\) approximate \(P(\mathbf{Z}|\mathbf{X})\). Since the Kullback-Leibler (KL) divergence \(\text{KL}(q(\mathbf{Z}|\mathbf{X}) \| P(\mathbf{Z}|\mathbf{X}))\) is non-negative and approaches zero as these two distributions become more similar, maximizing the term \(\int q(\mathbf{Z}|\mathbf{X}) \ln \left( \frac{P(\mathbf{X}, \mathbf{Z})}{q(\mathbf{Z}|\mathbf{X})} \right) d\mathbf{Z}\) minimizes the KL divergence. This term can be decomposed as:
    \begin{align}
    \int q(\mathbf{Z}|\mathbf{X}) \ln \left( \frac{P(\mathbf{X}, \mathbf{Z})}{q(\mathbf{Z}|\mathbf{X})} \right) d\mathbf{Z} &= \int q(\mathbf{Z}|\mathbf{X}) \ln P(\mathbf{X}|\mathbf{Z}) \, d\mathbf{Z} + \int q(\mathbf{Z}|\mathbf{X}) \ln \left( \frac{P(\mathbf{Z})}{q(\mathbf{Z}|\mathbf{X})} \right) d\mathbf{Z} \notag \\
    &= \mathbb{E}_{q(\mathbf{Z}|\mathbf{X})} [\ln P(\mathbf{X}|\mathbf{Z})] - \text{KL}(q(\mathbf{Z}|\mathbf{X}) \| P(\mathbf{Z})). \tag{5} 
    \end{align}
    
    Note the distinction between the KL divergence \(\text{KL}(q(\mathbf{Z}|\mathbf{X}) \| P(\mathbf{Z}))\) in (5) and \(\text{KL}(q(\mathbf{Z}|\mathbf{X}) \| P(\mathbf{Z}|\mathbf{X}))\) in (4). The expression in (5) is termed the Evidence Lower Bound (ELBO). From (4), maximizing the ELBO forces \(q(\mathbf{Z}|\mathbf{X})\) to approximate \(P(\mathbf{Z}|\mathbf{X})\). This is the principle of variational inference. To estimate \(P(\mathbf{Z}|\mathbf{X})\), it suffices to maximize the ELBO, making (5) the objective function for optimizing a Variational Autoencoder (VAE).
    
    \subsection{Why Variational Autoencoders are Considered Generative Models}
    The variational inference process outlined previously appears to focus solely on estimating the posterior \(P(\mathbf{Z}|\mathbf{X})\), seemingly unrelated to generation. However, maximizing the expectation \(\mathbb{E}_{q(\mathbf{Z}|\mathbf{X})} [\ln P(\mathbf{X}|\mathbf{Z})]\) corresponds to maximizing the expected log-likelihood of the observed data \(\mathbf{X}\) given the latent variables \(\mathbf{Z}\), where the expectation is taken over the variational distribution \(q(\mathbf{Z}|\mathbf{X})\). This process models the generation of \(\mathbf{X}\) from \(\mathbf{Z}\), with the generative process defined probabilistically.
    
    \subsection{VAE Framework for ICA: Modeling Latent Variables as Independent Components with Factorized Priors and Posteriors}
    In the Variational Autoencoder (VAE) framework, the encoding process from observed data \(\mathbf{X}\) to latent variables \(\mathbf{Z}\) is modeled by the variational distribution \(q(\mathbf{Z}|\mathbf{X})\), which represents the probability distribution of \(\mathbf{Z}\) given \(\mathbf{X}\). Conversely, the decoding process from \(\mathbf{Z}\) to \(\mathbf{X}\) is captured by \(P(\mathbf{X}|\mathbf{Z})\), denoting the probability of reconstructing \(\mathbf{X}\) from the latent variables \(\mathbf{Z}\). Thus, the VAE framework facilitates a mapping from observed input \(\mathbf{X}\) to latent variables \(\mathbf{Z}\) and back to a reconstructed \(\mathbf{X}\). Ideally, \(\mathbf{Z}\) should not merely serve as a compressed representation of \(\mathbf{X}\) but as variables with interpretable significance. If \(\mathbf{Z}\) can control or explain \(\mathbf{X}\), the VAE model achieves significant advancements in interpretability, causal inference, and disentanglement. This implies that the model captures independent factors influencing the properties of \(\mathbf{X}\), potentially expressing interpretable generative rules for \(\mathbf{X}\).
    
    Achieving structured, factorized, and independent dimensions of the latent variables \(\mathbf{Z} = \{\mathbf{Z}^a, \mathbf{Z}^b, \dots, \mathbf{Z}^n\}\) is critical to realizing this vision. As shown in Equation (5), maximizing the Evidence Lower Bound (ELBO) drives the Kullback-Leibler (KL) divergence \(\text{KL}(q(\mathbf{Z}|\mathbf{X}) \| P(\mathbf{Z}))\) toward zero, thereby aligning \(q(\mathbf{Z}|\mathbf{X})\) with the prior \(P(\mathbf{Z})\). To ensure that \(q(\mathbf{Z}|\mathbf{X})\) is factorized, independent, and structured, the prior \(P(\mathbf{Z})\) must be carefully designed to influence the posterior \(q(\mathbf{Z}|\mathbf{X})\) accordingly.
    
    In variational inference under the Bayesian framework, the prior distribution plays a pivotal role, as it is the only component that can be explicitly designed to influence inference outcomes. To achieve independence among the dimensions of the latent variables \(\mathbf{Z} = \{\mathbf{Z}^a, \mathbf{Z}^b, \dots, \mathbf{Z}^n\}\), the prior is supposed to satisfy the independence condition:
    \begin{equation}
    P(\mathbf{Z}^a, \mathbf{Z}^b, \dots, \mathbf{Z}^n) = P(\mathbf{Z}^a) \cdot P(\mathbf{Z}^b) \cdot \dots \cdot P(\mathbf{Z}^n). \tag{6} 
    \end{equation}
    This equality implies that the probability of any joint event can be computed as the product of the probabilities of the individual events, reflecting the absence of statistical dependence among the variables. As indicated by Equation (6), independence in probability theory implies factorization. Thus, the prior distribution is designed to be factorized as:
    \begin{equation}
    \ln P_{\Gamma}(\mathbf{Z}) = \sum_{i=a}^n \ln P_{\Gamma^i}(\mathbf{Z}^i). \tag{7} 
    \end{equation}
    Here, the prior is parameterized by \(\Gamma = \{\Gamma^a, \Gamma^b, \dots, \Gamma^n\}\), where each prior \(P_{\Gamma^i}(\mathbf{Z}^i)\) is governed by a distinct parameter \(\Gamma^i\). This parameterization encourages distinct independent prior distributions by allowing each \(\Gamma^i\) to differ.
    
    Similarly, the posterior distribution is factorized to align with the prior:
    \begin{equation}
    \ln q_{\Theta}(\mathbf{Z}|\mathbf{X}) = \sum_{i=a}^n \ln q_{\mathbf{\Theta}}(\mathbf{Z}^i|\mathbf{X}). \tag{8} 
    \end{equation}
    Through the Kullback-Leibler (KL) divergence, each posterior \(q_{\mathbf{\Theta}}(\mathbf{Z}^i|\mathbf{X})\) is driven to approximate its corresponding prior \(P_{\Gamma^i}(\mathbf{Z}^i)\), where the prior’s distribution is shaped by \(\Gamma^i\). Consequently, the posterior distributions can be made mutually independent by designing the prior parameters \(\Gamma\). Typically, these parameters are set as trainable and optimized through the VAE objective function, which naturally drives \(\Gamma^a, \Gamma^b, \dots, \Gamma^n\) toward distinct values. Additional mechanisms can be introduced to further encourage diversity among \(\Gamma^i\) if needed.
    
    With this design, the VAE encoding process, which yields the posterior \(q_{\mathbf{\Theta}}(\mathbf{Z}^i|\mathbf{X})\), is equivalent to the inverse problem in Independent Component Analysis (ICA), where observed signals \(\mathbf{X}\) are mapped to independent components or sources \(\mathbf{Z}^i\). This approach, which emphasizes structured, factorized, independentdent and disentangled latent dimensions, distinguishes the proposed framework from traditional VAEs.
    
    \subsection{Enhancing Latent Variable Independence with Adversarial Networks}
    
    The previous section demonstrated how structured and factorized prior distributions can promote independence among the dimensions of latent variables \(\mathbf{Z} = \{\mathbf{Z}^a, \mathbf{Z}^b, \dots, \mathbf{Z}^n\}\). This section introduces adversarial networks as a mechanism to further encourage mutual independence among these dimensions.
    
    Adversarial networks are valuable for implicitly aligning two distributions (\cite{ganin2016domain}; \cite{arjovsky2017wasserstein}; \cite{tzeng2017adversarial}; \cite{liang2021well}). By sampling from two distributions and training a discriminator to distinguish between them, the distributions are deemed similar when the discriminator cannot differentiate the samples. This approach does not require explicit knowledge of the distributions’ analytical forms, only access to their samples, enabling one distribution to approximate another or both to approach mutually.
    
    This property is leveraged to promote independence among latent variable dimensions, as defined in Equation (6):
    \begin{equation}
    R(\mathbf{Z}^a, \mathbf{Z}^b, \dots, \mathbf{Z}^n) = R(\mathbf{Z}^a) \cdot R(\mathbf{Z}^b) \cdot \dots \cdot R(\mathbf{Z}^n). \tag{6} 
    \end{equation}
    If samples from the joint distribution and marginal distributions of the latent dimensions can be obtained and fed into a discriminator to minimize their differences, the dimensions can be made as independent as possible. \cite{brakel2017learning} proposed a method to derive marginal distribution samples from joint distribution samples. Each dimension is represented as a sequence \(\mathbf{Z}_{1:T}^i\), where the \(T\) points form the marginal distribution \(R(\mathbf{Z}_{1:T}^i)\). Samples with consistent indices across \(n\) dimensions, i.e., \(\{Z_\tau^a, Z_\tau^b, \dots, Z_\tau^n\}\), represent the joint distribution, while samples with mismatched indices, i.e., \(\{Z_{?}^a, Z_{?}^b, \dots, Z_{?}^n\}\), represent the marginal distributions. The most effective method to obtain mismatched samples is to shuffle the sequence \({1:T}\), preserving each marginal distribution \(R(\mathbf{Z}^i)\) (changing the sequence order of \(\mathbf{Z}_{1:T}^i\) does not change its marginal distribution). This shuffling strategy, detailed in (\cite{wei2024innovative}), is more intuitive than the resampling approach proposed by \cite{brakel2017learning}, though both yield comparable results.
     
    The mechanism of using adversarial networks to promote latent dimension independence is illustrated in Figure 1, and the integration of this module into the VAE architecture is shown in Figure 2. Strictly, the sequence \(\mathbf{Z}^i_{1:T}\) corresponds to \(q_{\mathbf{\Theta}}(\mathbf{Z}^i|\mathbf{X})\), where each point \({Z}^i_\tau\) is a distribution parameterized by mean \(\mu_{{Z}^i_\tau}\) and variance \(\mathbf{\sigma}^{2}_{{Z}^{i}_\tau}\). For simplicity, a single variance \(\mathbf{\sigma}^{2}_{{Z}^{i}}\) can be shared across all points in the sequence \(\mathbf{\mu}_{\mathbf{Z}^i_{1:T}}\), reducing the number of parameters (\cite{wei2024innovative}). In the adversarial module, the sequence \(\mathbf{\mu}_{\mathbf{Z}^i_{1:T}}\) represents \(\mathbf{Z}_{1:T}^i\). Note that the marginal distribution \(R(\mathbf{Z}^i)\) refers to the distribution of the mean sequence \(\mathbf{\mu}_{\mathbf{Z}^i_{1:T}}\), distinct from \(q_{\mathbf{\Theta}_i}(\mathbf{Z}^i|\mathbf{X})\), which is a multivariate Gaussian distribution over the sequence points.
    \begin{figure}[ht]
        \centering
        \setlength{\abovecaptionskip}{1pt}
        \includegraphics[width=\textwidth]{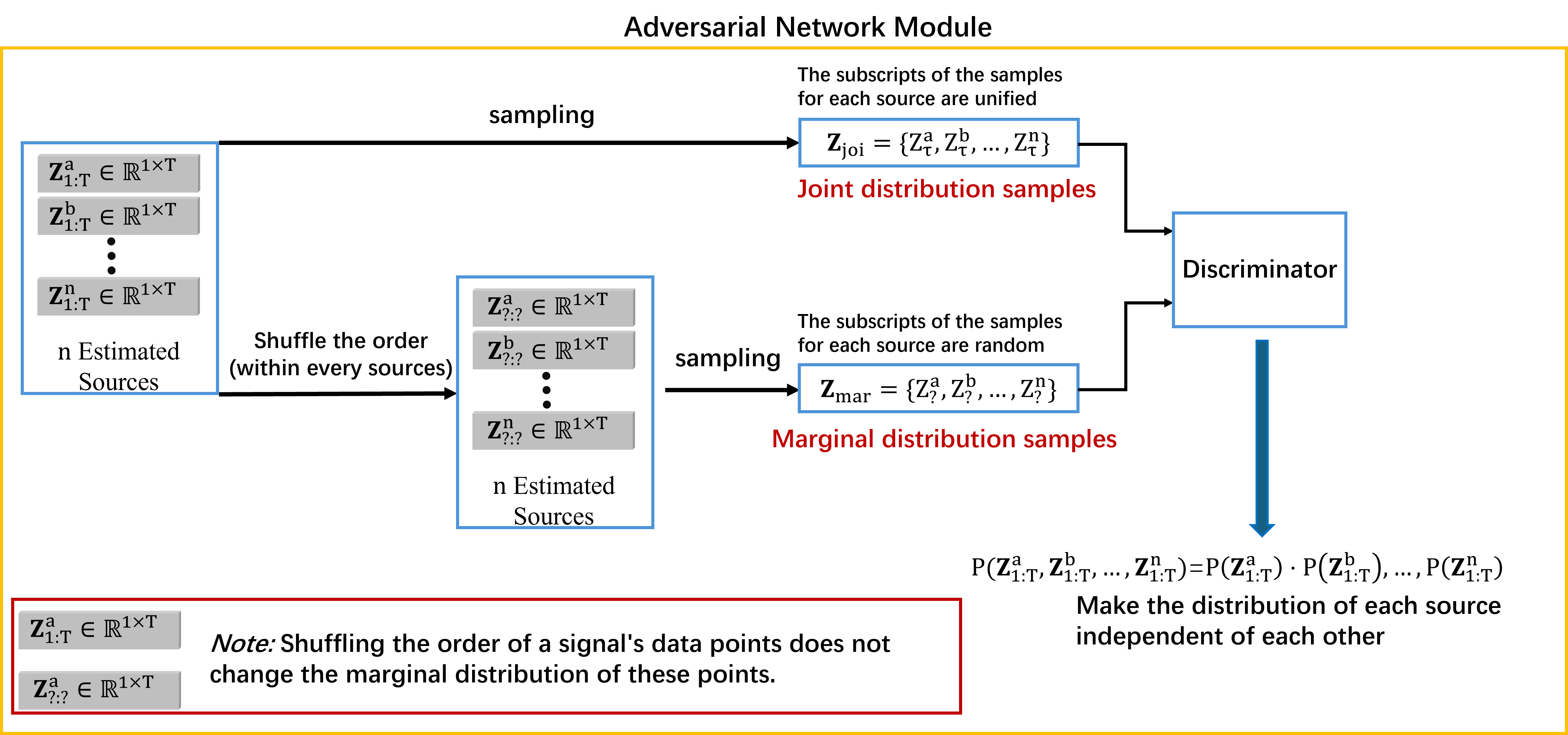}
        \caption{Mechanism of Using Adversarial Networks to Promote Latent Dimension Independence}
        \label{Figure1}
    \end{figure}
    \begin{figure}[ht]
        \centering
        \setlength{\abovecaptionskip}{1pt}
        \includegraphics[width=\textwidth]{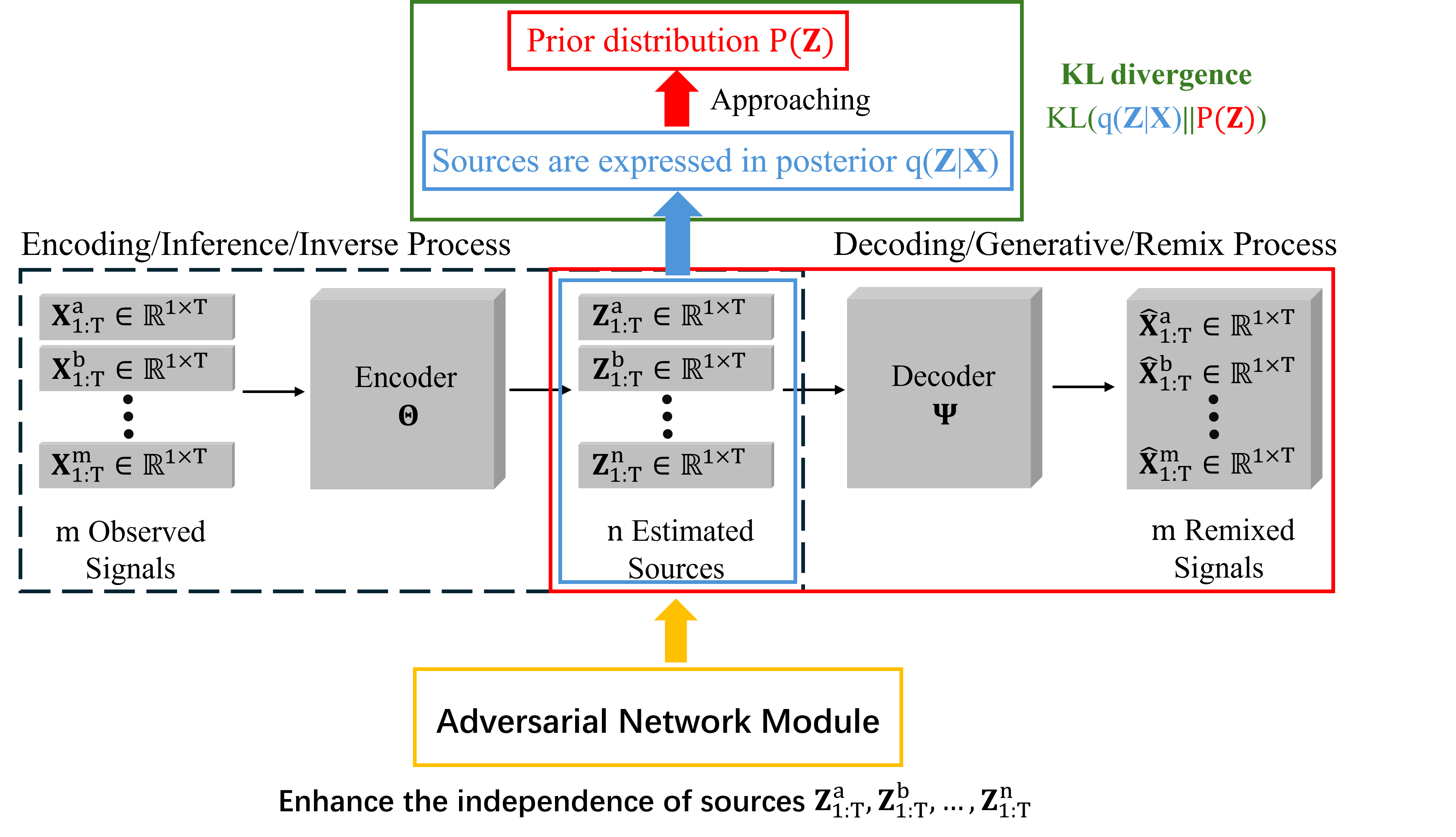}
        \caption{Adversarial Network Module Integrated with VAE Architecture for Enhancing Independence Among Latent Variable Dimensions}
        \label{Figure2}
    \end{figure}	
    The loss function for the adversarial network is defined as (\cite{goodfellow2014generative}):
    \begin{equation}
    L_D(\mathbf{\Phi}| \mathbf{Z}_{\text{mar}}, \mathbf{Z}_{\text{joi}}) = -\mathbb{E}_{\tilde{\mathbf{Z}}_{\text{mar}} \sim \prod_{i=1}^n R(Z^i)} [\log D_\mathbf{\Phi}(\tilde{\mathbf{Z}}_{\text{mar}})] - \mathbb{E}_{\tilde{\mathbf{Z}}_{\text{joi}} \sim R(\mathbf{Z})} [\log(1 - D_\mathbf{\Phi}(\tilde{\mathbf{Z}}_{\text{joi}}))]. \tag{9} 
    \end{equation}
    For samples from the joint distribution, we let:
    \begin{equation}
    \tilde{\mathbf{Z}}_{\text{joi}} = \{Z_\tau^a, Z_\tau^b, \dots, Z_\tau^n\} \approx  \{\mu_{Z_\tau}^a, \mu_{Z_\tau}^b, \dots, \mu_{Z_\tau}^n\} \sim R(\mathbf{Z}),\tag{10} 
    \end{equation} 
    and for samples from the marginal distribution, we have:
    
    \begin{equation}
    \tilde{\mathbf{Z}}_{\text{mar}} = \{Z_{?}^a, Z_{?}^b, \dots, Z_{?}^n\} \approx \{\mu_{Z_{?}}^a, \mu_{Z_{?}}^b, \dots, \mu_{Z_{?}}^n\} \sim \prod_{i=1}^n R(Z^i). \tag{11} 
    \end{equation} 
    
    Here, \(\mathbf{\Phi}\) denotes the discriminator parameters. These samples are input to the discriminator \(D_\Phi\), which, through adversarial training, drives the marginal distribution product \(\prod_{i=1}^n R(Z^i)\) to approximate the joint distribution \(R(\mathbf{Z})\).
    
    \subsection{Encoder-Free Strategy: Bypass Explicit Inverse Mapping}
    
    As illustrated in Figure 2, the variational distribution \(q_{\mathbf{\Theta}}(\mathbf{Z}|\mathbf{X})\) in the objective function Equation (5), representing the mapping from observed data \(\mathbf{X}\) to latent variables \(\mathbf{Z}\), i.e., \(\mathbf{Z} = f^{-1}(\mathbf{X})\), is implemented by the encoder. The encoder effectively performs the inverse mapping \(f^{-1}\). However, in underdetermined conditions, where the dimensionality of \(\mathbf{X}\) (\(m\)) is less than that of \(\mathbf{Z}\) (\(n\)), the inverse mapping \(f^{-1}\) does not exist, rendering explicit representation via the encoder infeasible (\cite{wei2024half}). Here, the encoder, typically a multilayer perceptron or neural network parameterized by \(\mathbf{\Theta}\), cannot explicitly execute \(f^{-1}\).
    
    To address this, our prior work introduced the Half-VAE, which eliminates the encoder from the VAE framework (\cite{wei2024half}). This approach avoids explicitly defining the inverse mapping \(f^{-1}\) through the encoding process. In a standard VAE, the encoder parameters \(\mathbf{\Theta}\) are optimized via the objective function, with latent variables \(\mathbf{Z}\) as the mapping output. In contrast, Half-VAE treats the latent variables \(\mathbf{Z}\) as directly optimizable parameters due to the absence of an encoder. The loss functions for standard VAE and Half-VAE, shown below, highlight the distinction in optimized parameters. To align with neural network optimization via stochastic gradient descent, maximizing the Evidence Lower Bound (ELBO) is reformulated as minimizing its negative, termed the loss function:
    \begin{equation}
    L_{\text{VAE}}(\mathbf{\Theta}, \mathbf{\Psi}, \mathbf{\Gamma} | \mathbf{X}) = -\mathbb{E}_{q_{\Theta}(\mathbf{Z}|\mathbf{X})} [\ln P_{\mathbf{\Psi}}(\mathbf{X}|\mathbf{Z})] + \text{KL}(q_{\mathbf{\Theta}}(\mathbf{Z}|\mathbf{X}) \| P_{\mathbf{\Gamma}}(\mathbf{Z})). \tag{12} 
    \end{equation}
    \begin{equation}
    L_{\text{Half-VAE}}(\mathbf{\Omega}, \mathbf{\Psi}, \mathbf{\Gamma} | \mathbf{X}) = -\mathbb{E}_{I_{\mathbf{\Omega}}(\mathbf{Z})} [\ln P_{\mathbf{\Psi}}(\mathbf{X}|\mathbf{Z})] + \text{KL}(I_{\mathbf{\Omega}}(\mathbf{Z}) \| P_{\mathbf{\Gamma}}(\mathbf{Z})). \tag{13}
    \end{equation}
    Here, \(\mathbf{\Psi}\) denotes the decoder parameters, and \(\mathbf{\Omega} = (\mathbf{\mu_z}, \mathbf{\sigma_z^2)}\) represents the distribution parameters of \(I_{\mathbf{\Omega}}(\mathbf{Z})\), optimized directly through Equation (13). Previous work (\cite{wei2024half}) demonstrated that the posterior approximation \(q_{\mathbf{\Theta}}(\mathbf{Z}|\mathbf{X})\) is not the only viable form in variational inference and that the encoder and its explicit mapping are not essential components of a VAE. Thus, Half-VAE replaces \(\text{KL}(q(\mathbf{Z}|\mathbf{X}) \| P(\mathbf{Z}|\mathbf{X}))\) in Equation (4) with \(\text{KL}(I_{\mathbf{\Omega}}(\mathbf{Z}) \| P(\mathbf{Z}|\mathbf{X}))\), using \(I_{\mathbf{\Omega}}(\mathbf{Z})\) to directly approximate the true posterior \(P(\mathbf{Z}|\mathbf{X})\) instead of relying on the encoder’s output \(q(\mathbf{Z}|\mathbf{X})\).
    
    \subsection{Half Adversarial VAE}
    
    Building on the structured prior design of Variational Autoencoders (VAEs), the use of adversarial networks to promote factorization and independence of latent variable dimensions, and the encoder-free approach of Half-VAE, this study proposes the Half Adversarial VAE (Half-AVAE). A schematic of Half-AVAE is shown in Figure 3. Compared to Half-VAE, Half-AVAE incorporates an adversarial network to further encourage mutual independence among the latent dimensions \(\mathbf{Z} = \{\mathbf{Z}^a, \mathbf{Z}^b, \dots, \mathbf{Z}^n\}\). The Half-AVAE framework is governed by two loss functions:
    \begin{equation}
    L_{\text{Half-AVAE}_1}(\mathbf{\Omega}, \mathbf{\Psi}, \mathbf{\Gamma} | \mathbf{X}) = -\mathbb{E}_{I_{\mathbf{\Omega}}(\mathbf{Z})} [\ln P_{\mathbf{\Psi}}(\mathbf{X}|\mathbf{Z})] + \text{KL}(I_{\mathbf{\Omega}}(\mathbf{Z}) \| P_{\Gamma}(\mathbf{Z})) - \Lambda \cdot L_{\text{Half-AVAE}_2}. \tag{14a} 
    \end{equation}
    \begin{equation}
    L_{\text{Half-AVAE}_2}(\mathbf{\Phi} | \mathbf{Z}_{\text{mar}}, \mathbf{Z}_{\text{joi}}) = -\mathbb{E}_{\tilde{\mathbf{Z}}_{\text{mar}} \sim \prod_{i=1}^n R(Z^i)} [\log D_{\mathbf{\Phi}}(\tilde{\mathbf{Z}}_{\text{mar}})] - \mathbb{E}_{\tilde{\mathbf{Z}}_{\text{joi}} \sim R(\mathbf{Z})} [\log(1 - D_{\mathbf{\Phi}}(\tilde{\mathbf{Z}}_{\text{joi}}))]. \tag{14b} 
    \end{equation}
    The loss function \(L_{\text{Half-AVAE}_1}\) optimizes the Half-VAE parameters \(\mathbf{\Omega}\), \(\mathbf{\Psi}\), and \(\mathbf{\Gamma}\), where \(\mathbf{\Omega} = (\mathbf{\mu}_z, \mathbf{\sigma}_z^2)\) defines the distribution \(I_{\mathbf{\Omega}}(\mathbf{Z})\), \(\mathbf{\Psi}\) denotes the decoder parameters, and \(\mathbf{\Gamma}\) parameterizes the prior \(P_{\mathbf{\Gamma}}(\mathbf{Z})\). The loss function \(L_{\text{Half-AVAE}_2}\) optimizes the adversarial network parameters \(\mathbf{\Phi}\). The hyperparameter \(\Lambda\) modulates the influence of the adversarial loss on the Half-VAE, encouraging the inference of mutually independent latent dimensions to compete with the discriminator \(D_{\mathbf{\Phi}}\) (\cite{brakel2017learning}).
    \begin{figure}[ht]
        \centering
        \setlength{\abovecaptionskip}{1pt}
        \includegraphics[width=\textwidth]{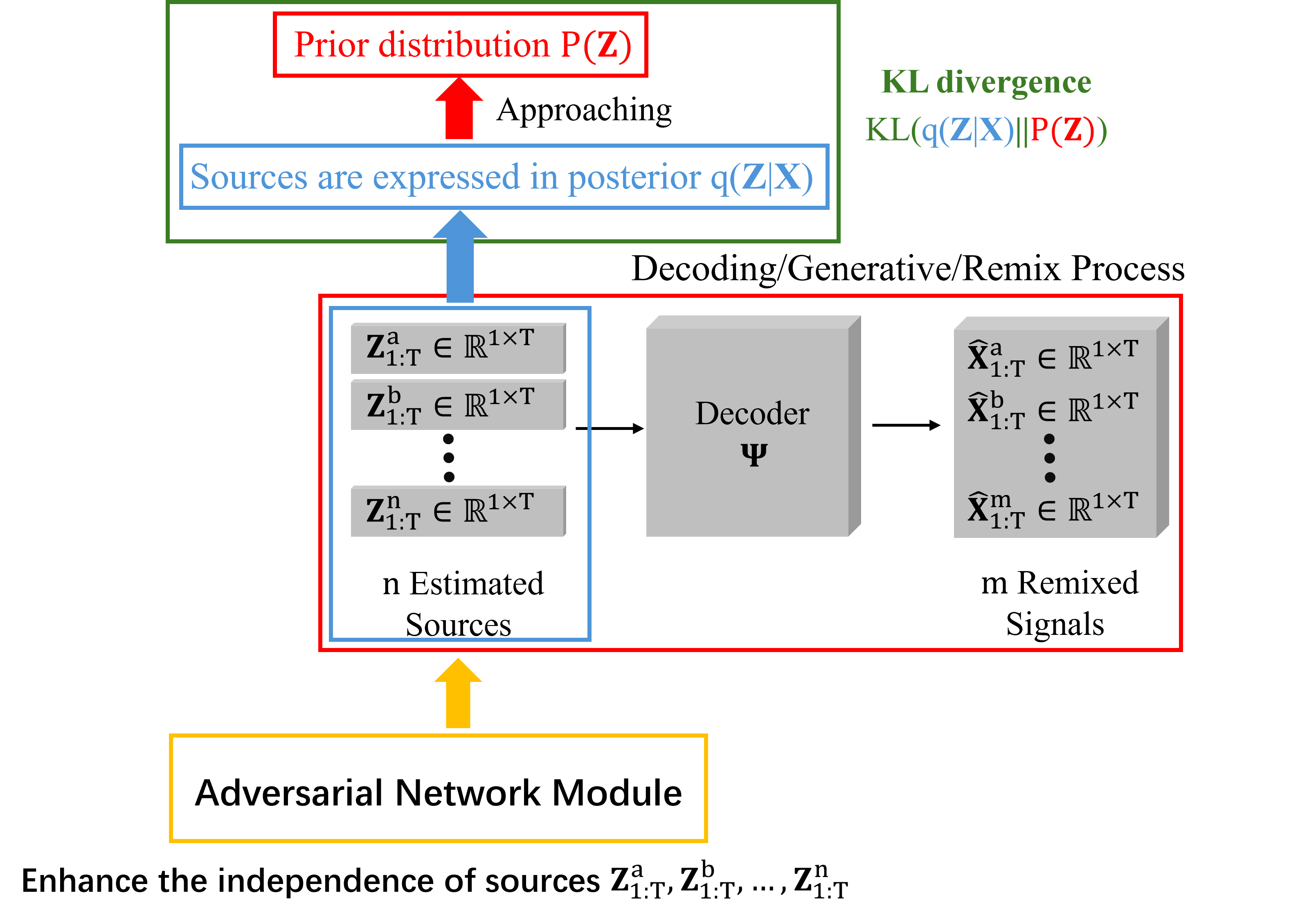}
        \caption{Half-AVAE}
        \label{Figure3}
    \end{figure}
    
    \section{Experiments}
    
    In my prior work (\cite{wei2024innovative}), each dimension of the latent variables was assigned an independent Gaussian Process (GP) prior, enabling the posterior \(q_{\mathbf{\Theta}}(\mathbf{Z}^i|\mathbf{X})\) to approximate its respective GP prior, thus endowing each dimension with distinct temporal or spatial structures. To accommodate potential independent and identically distributed (i.i.d.) signals, an adversarial network was introduced to enhance the independence among the mean sequences of the latent dimensions. This approach, termed Gaussian Process Adversarial Variational Autoencoder (GP-AVAE), is the first to combine GP priors and adversarial networks for Independent Component Analysis (ICA) or disentanglement tasks within the VAE framework.
    
    In the earlier Half-VAE framework (\cite{wei2024half}), the priors for each latent dimension were independent Gaussian Mixture Models (GMMs). Since complex distributions can theoretically be approximated by linear combinations of Gaussians, GMM priors offer flexibility in modeling diverse latent variable distributions.
    
    \subsection{Synthetic Data}
    
    In this study, three source signal sequences with distinct temporal or spatial structures (i.e., varying correlations between points) were generated, as illustrated in Figure 4. To capture these structures, the prior for each latent dimension, \(P_{\Gamma^i}(\mathbf{Z}^i_{1:T})\), is designed as a Gaussian Process (GP), with each GP defined by its kernel function (covariance matrix). The commonly used Squared Exponential (SE) kernel was adopted (\cite{wei2024innovative}):
    \begin{equation}
    k_i(\tau, \tau') = \exp\left(-\frac{1}{2{\Gamma^i}^2}(\tau - \tau')^2\right). \tag{15} 
    \end{equation}
    Here, the parameter \(\Gamma^i\) represents the length scale of the GP prior \(P_{\Gamma^i}(\mathbf{Z}^i_{1:T})\), modeling the strength of correlation between points \(\tau\) and \(\tau'\) in a sequence. If alternative prior distributions are used, such as GMM in (\cite{wei2024half}), \(\Gamma^i\) would denote the parameters characterizing those distributions.

    \begin{figure}[ht]
        \centering
        \setlength{\abovecaptionskip}{1pt}
        \includegraphics[width=\textwidth]{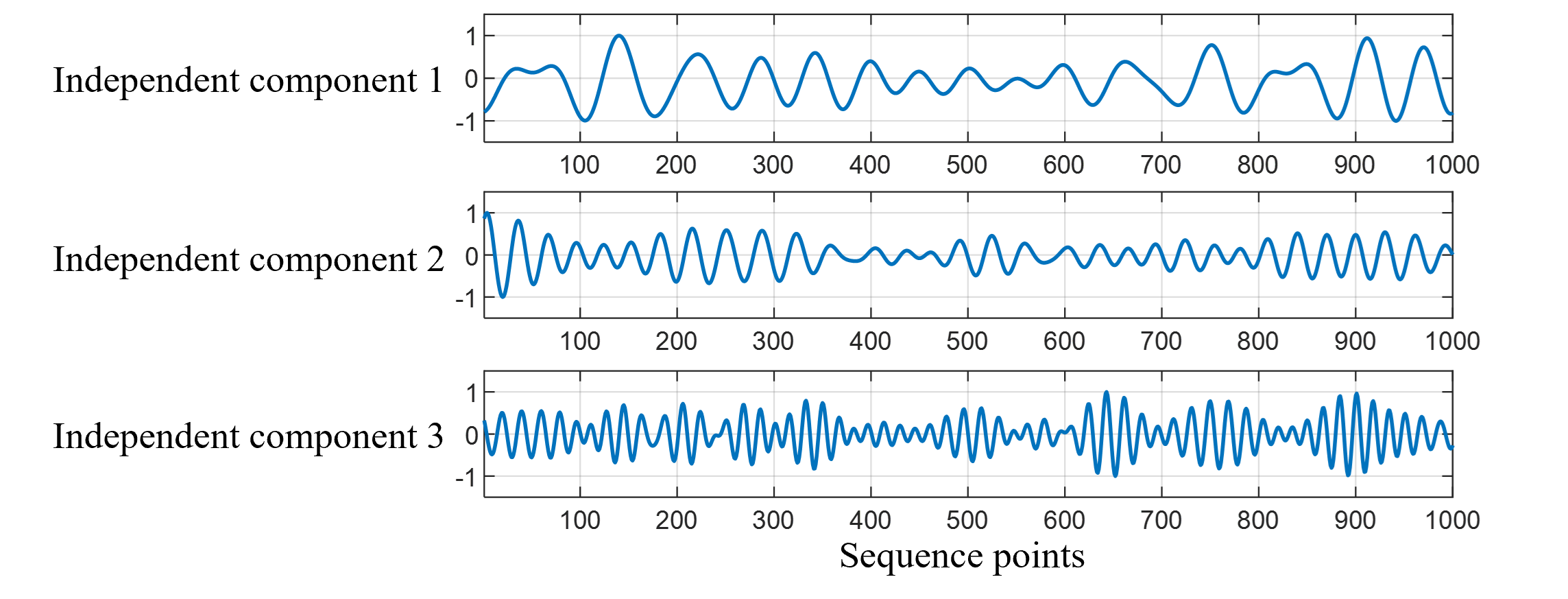}
        \caption{Synthetic Sources}
        \label{Figure4}
    \end{figure}
    
    \subsection{Determined Scenario}
    
    This study first evaluates the performance of Half-GP-AVAE in solving Independent Component Analysis (ICA) under determined conditions, where the number of observed signals \(\mathbf{X}\) equals the number of source signals, i.e., \(m = n\). This evaluation is a standard procedure, as failure to address the determined case would undermine further investigation. Given the use of GP priors, Half-AVAE is equivalently denoted as Half-GP-AVAE. For comparison, GP-AVAE and Half-VAE (also referred to as Half-GP-VAE) are employed as baseline methods to assess the results of Half-GP-AVAE.
    
    Figure 5 illustrates the observations \(\mathbf{X}\), obtained via the mixing mapping \(f\), i.e., \(\mathbf{X} = f(\mathbf{Z})\). Figure 6 presents the ICA results for GP-AVAE, Half-GP-VAE, and Half-GP-AVAE under determined conditions. The ground truth independent components are depicted as solid blue lines, while the inferred results from each method are shown as red dashed lines. Due to the scale ambiguity inherent in ICA, all signals are normalized using z-scores for consistent comparison. Notably, the inferred results from GP-AVAE, Half-GP-VAE, and Half-GP-AVAE are expressed probabilistically. Each inferred independent component sequence  \(\mathbf{Z}^i_{1:T}\)  is characterized by a mean sequence \(\mathbf{\mu}_{\mathbf{Z}^i_{1:T}}\)  and a single variance parameter \(\mathbf{\sigma}^{2}_{{Z}^{i}}\), where all points in a sequence share the same variance, but different sequences have distinct variances. For simplicity, the red dashed lines in Figure 6 only represent the mean sequences \(\mathbf{\mu}_{\mathbf{Z}^i_{1:T}}\).
    
    Under this determined condition, GP-AVAE, Half-GP-VAE, and Half-GP-AVAE accurately recover the independent components.  Half-GP-AVAE does not exhibit a clear advantage over other baseline methods.

    \begin{figure}[ht]
        \centering
        \setlength{\abovecaptionskip}{1pt}
        \includegraphics[width=\textwidth]{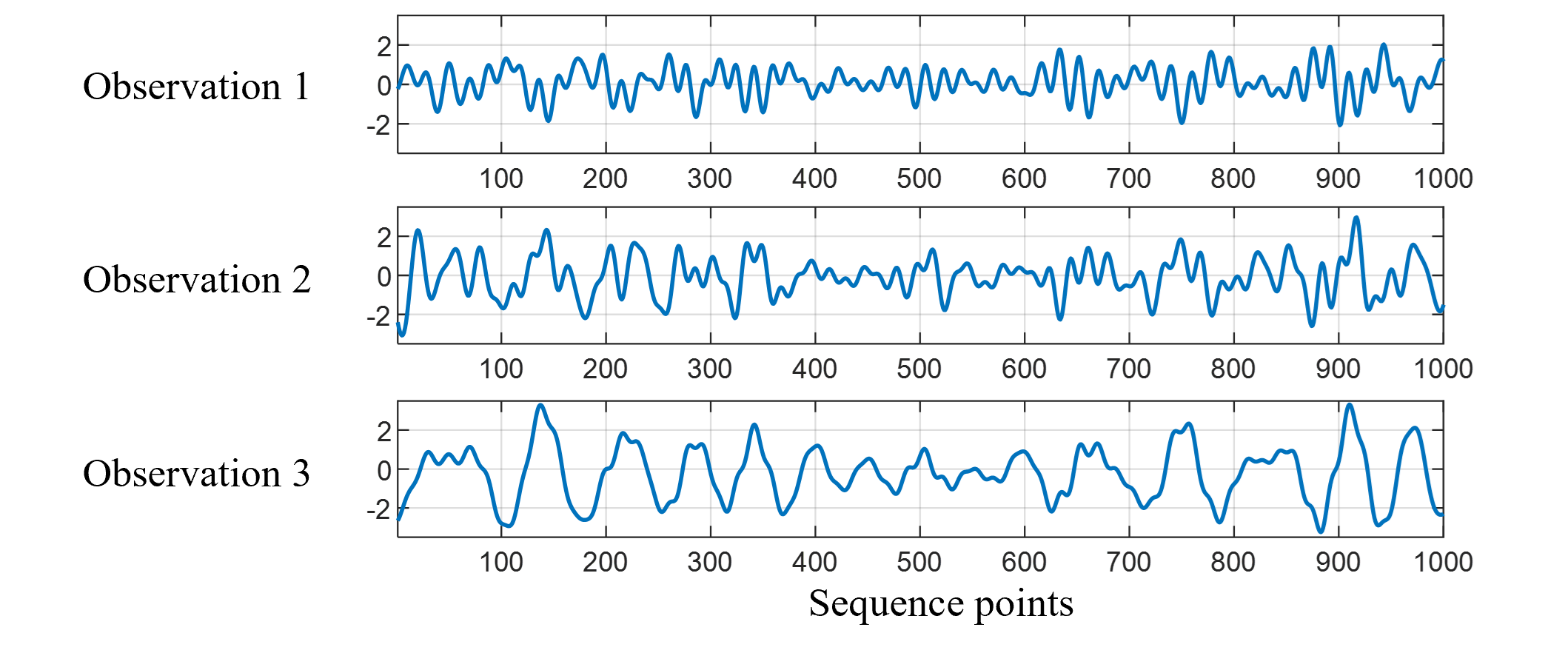}
        \caption{Observations of Determined Scenario}
        \label{Figure5}
    \end{figure}
    \begin{figure}[ht]
        \centering
        \setlength{\abovecaptionskip}{1pt}
        \includegraphics[width=\textwidth]{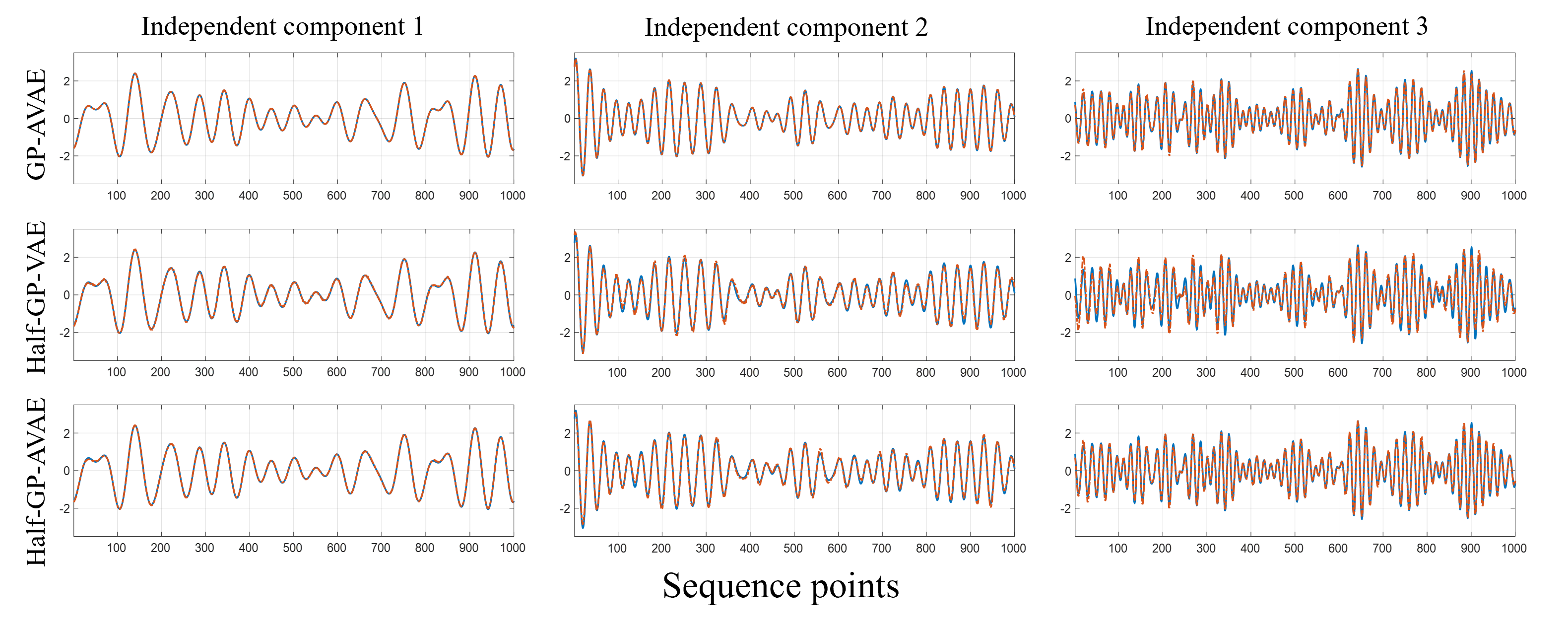}
        \caption{ICA Results of Determined Scenario}
        \label{Figure6}
    \end{figure}
    
    \subsection{Underdetermined Scenario}
    \subsubsection{Scenario without Additional Independence Mechanisms}
    
    Figure 7 illustrates the observed signals \(\mathbf{X}\) under underdetermined conditions, where the number of observed signals is fewer than the source signals. Figure 8 presents the performance of different methods, with the root mean square error (RMSE) between the inferred independent components and their corresponding ground truth, under z-score normalization, reported in Table 1. The results indicate that Half-GP-VAE and Half-GP-AVAE slightly outperform GP-AVAE, though no significant advantage is observed.

    \begin{figure}[ht]
        \centering
        \setlength{\abovecaptionskip}{1pt}
        \includegraphics[width=\textwidth]{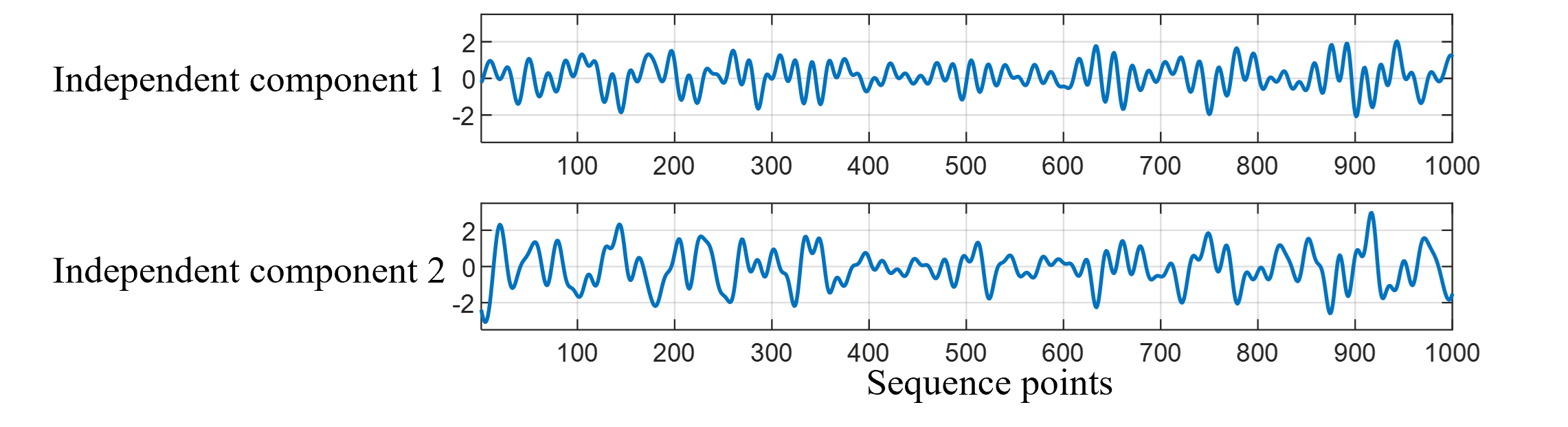}
        \caption{Observations of Underdetermined Scenario}
        \label{Figure7}
    \end{figure}
    \begin{figure}[ht]
        \centering
        \setlength{\abovecaptionskip}{1pt}
        \includegraphics[width=\textwidth]{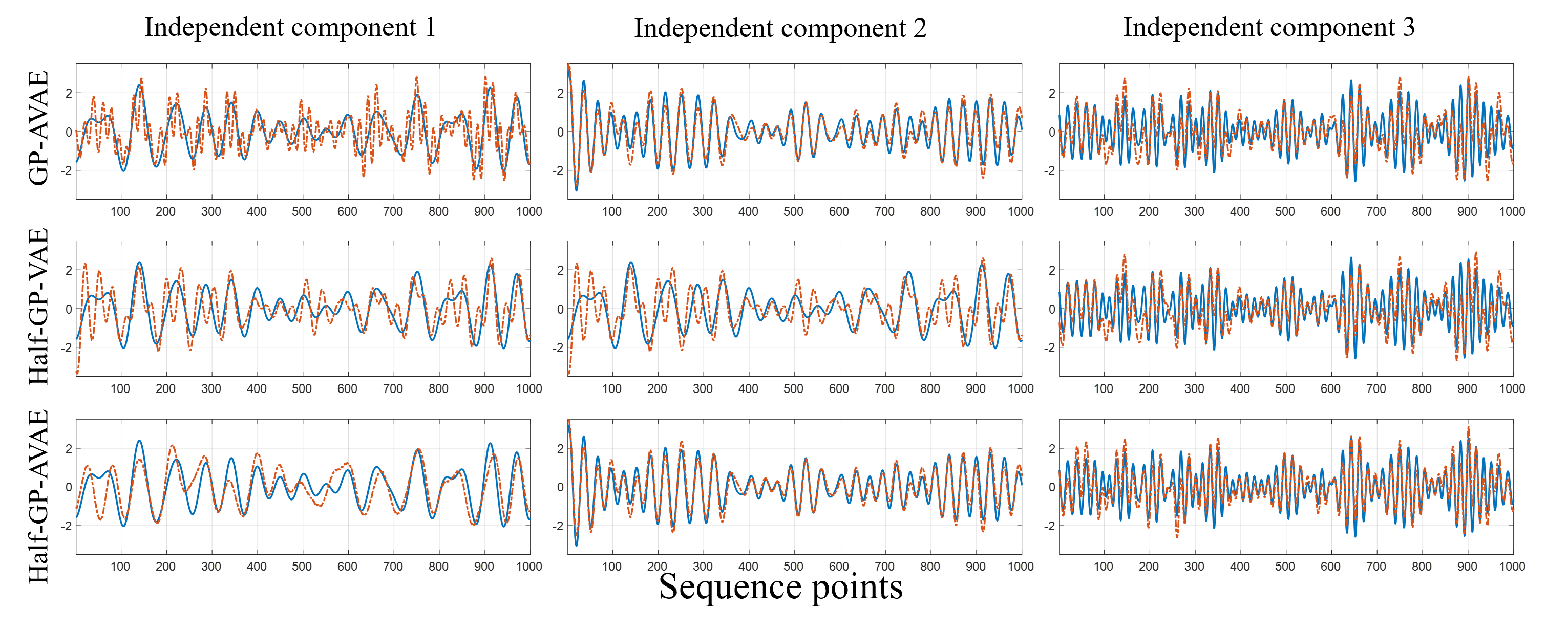}
        \caption{ICA Results of Underdetermined Scenario}
        \label{Figure8}
    \end{figure}

    \begin{table}
    \centering
    \begin{tabular}{l|ccc}
     & GP-AVAE & Half-GP-VAE & Half-GP-AVAE \\\hline
    Independent component 1 & 0.8494 & 0.8468 & 0.6981 \\
    Independent component 2 & 0.4021 & 0.1736 & 0.3317 \\
    Independent component 3 & 0.6941 & 0.7256 & 0.5185 \\
    Average & 0.6485 & 0.582 & 0.5161 \\
    \end{tabular}
    \caption{\label{tab:results1}ICA Performance}
    \end{table}
    
    \subsubsection{Enhancing Independence and Factorization}
    
    To address the insufficient independence and factorization of latent variable dimensions in Half-GP-AVAE’s adversarial network, External Enhancement (EE) terms are introduced to promote mutual independence among the dimensions \(\mathbf{Z} = \{\mathbf{Z}^a, \mathbf{Z}^b, \dots, \mathbf{Z}^n\}\). The EE terms are defined as:
    \begin{equation}
    \text{EE} = \beta_1 \sum_{1 \leq i < j \leq n} \frac{1}{(\Gamma^i - \Gamma^j)^2} + \beta_2 \sum_{i=1}^n \Gamma^i + \beta_3 \sum_{i=1}^n \mathbf{\sigma}^{2}_{{Z}^{i}}. \tag{16}
    \end{equation}
    The first term encourages the prior distribution parameters \(\Gamma^i\) to diverge, maximizing the differences between the priors of each latent dimension. During model optimization, \(\Gamma^i\) and \(\mathbf{\sigma}^{2}_{{Z}^{i}}\) may grow unbounded with increasing epochs, necessitating the second and third terms to constrain their values. The hyperparameters \(\beta_1\), \(\beta_2\), and \(\beta_3\) are tuned to balance the absolute magnitudes of these terms during convergence.
    
    For fair comparison, the EE terms are incorporated into the loss functions of GP-AVAE, Half-GP-VAE, and Half-GP-AVAE:
    \begin{align}
    L_{\text{GP-AVAE}_1}(\mathbf{\Theta}, \mathbf{\Psi}, \mathbf{\Gamma} | \mathbf{X}) &= -\mathbb{E}_{q_{\mathbf{\Theta}}(\mathbf{Z}|\mathbf{X})} [\ln P_{\mathbf{\Psi}}(\mathbf{X}|\mathbf{Z})] + \text{KL}(q_{\mathbf{\Theta}}(\mathbf{Z}|\mathbf{X}) \| \text{GP}_{\mathbf{\Gamma}}(\mathbf{Z})) - \Lambda \cdot L_{\text{GP-AVAE}_2} + \text{EE}, \tag{17a}
    \end{align}
    \begin{align}
    L_{\text{GP-AVAE}_2}(\mathbf{\Phi} | \mathbf{Z}_{\text{mar}}, \mathbf{Z}_{\text{joi}}) &= -\mathbb{E}_{\tilde{\mathbf{Z}}_{\text{mar}} \sim \prod_{i=1}^n R(Z^i)} [\log D_{\mathbf{\Phi}}(\tilde{\mathbf{Z}}_{\text{mar}})] - \mathbb{E}_{\tilde{\mathbf{Z}}_{\text{joi}} \sim R(\mathbf{Z})} [\log(1 - D_{\mathbf{\Phi}}(\tilde{\mathbf{Z}}_{\text{joi}}))], \tag{17b}
    \end{align}
    \begin{align}
    L_{\text{Half-GP-VAE}}(\mathbf{\Omega}, \mathbf{\Psi}, \mathbf{\Gamma} | \mathbf{X}) = -\mathbb{E}_{I_{\mathbf{\Omega}}(\mathbf{Z})} [\ln P_{\mathbf{\Psi}}(\mathbf{X}|\mathbf{Z})] + \text{KL}(I_{\mathbf{\Omega}}(\mathbf{Z}) \| \text{GP}_{\mathbf{\Gamma}}(\mathbf{Z})) + \text{EE}, \tag{18}
    \end{align}
    \begin{align}
    L_{\text{Half-GP-AVAE}_1}(\mathbf{\Omega}, \mathbf{\Psi}, \mathbf{\Gamma} | \mathbf{X}) &= -\mathbb{E}_{I_{\mathbf{\Omega}}(\mathbf{Z})} [\ln P_{\mathbf{\Psi}}(\mathbf{X}|\mathbf{Z})] + \text{KL}(I_{\mathbf{\Omega}}(\mathbf{Z}) \| \text{GP}_{\mathbf{\Gamma}}(\mathbf{Z})) - \mathbf{\Lambda} \cdot L_{\text{Half-GP-AVAE}_2} + \text{EE}, \tag{19a}
    \end{align}
    \begin{align}
    L_{\text{Half-GP-AVAE}_2}(\mathbf{\Phi} | \mathbf{Z}_{\text{mar}}, \mathbf{Z}_{\text{joi}}) &= -\mathbb{E}_{\tilde{\mathbf{Z}}_{\text{mar}} \sim \prod_{i=1}^n R(Z^i)} [\log D_{\Phi}(\tilde{\mathbf{Z}}_{\text{mar}})] - \mathbb{E}_{\tilde{\mathbf{Z}}_{\text{joi}} \sim R(\mathbf{Z})} [\log(1 - D_{\Phi}(\tilde{\mathbf{Z}}_{\text{joi}}))]. \tag{19b} 
    \end{align}
    The prior \(P_{\Gamma}(\mathbf{Z})\) is denoted as \(\text{GP}_{\Gamma}(\mathbf{Z})\) to reflect the use of Gaussian Process priors in this study.
    
    Figure 9 illustrates the performance of GP-AVAE, Half-GP-VAE, and Half-GP-AVAE after incorporating the EE terms. The root mean square error (RMSE) between the inferred independent components and their ground truth, under z-score normalization, is reported in Table 2. The EE terms yield minimal improvement in GP-AVAE’s ICA performance, as its encoder cannot effectively address the inverse mapping under underdetermined conditions. Half-GP-VAE shows moderate improvement with the EE terms, suggesting that further enhancements may be needed for underdetermined ICA tasks. In contrast, Half-GP-AVAE, augmented by both the adversarial network and EE terms, achieves significantly reduced differences between inferred components and ground truth, effectively addressing the underdetermined ICA task.

    \begin{figure}[ht]
        \centering
        \setlength{\abovecaptionskip}{1pt}
        \includegraphics[width=\textwidth]{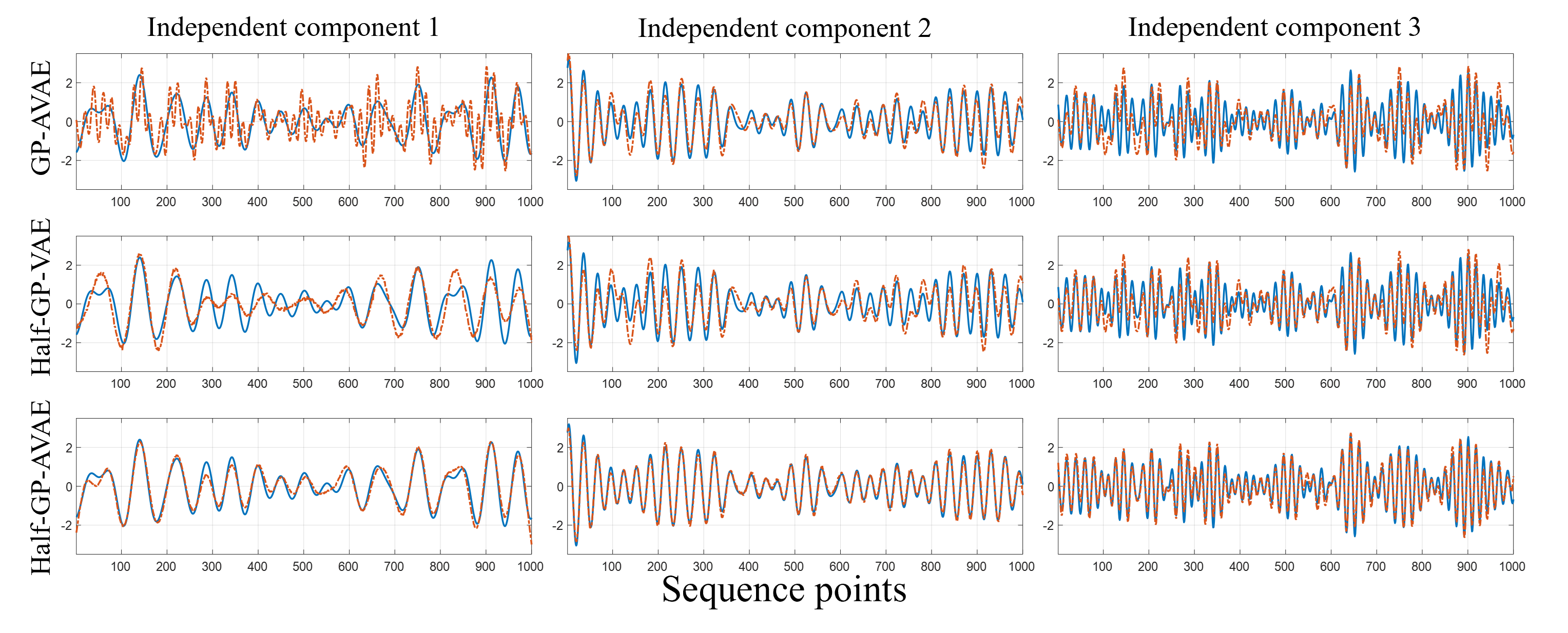}
        \caption{ICA Results of Underdetermined Scenario}
        \label{Figure9}
    \end{figure}

    \begin{table}
        \centering
        \begin{tabular}{l|ccc}
         & GP-AVAE & Half-GP-VAE & Half-GP-AVAE \\\hline
        Source 1 & 0.8488 & 0.5716 & 0.2653 \\
        Source 2 & 0.3994 & 0.5699 & 0.1449 \\
        Source 3 & 0.694 & 0.5854 & 0.2716 \\
        Average & 0.6474 & 0.5756 & 0.2272 \\
        \end{tabular}
        \caption{\label{tab:results2}Performance metrics across different sources.}
    \end{table}
    
    \section{Discussion and Conclusion}
    Theoretically, Variational Autoencoders (VAEs) with explicit encoders cannot perform the inverse mapping required for Independent Component Analysis (ICA) under underdetermined conditions, where the number of observed signals is fewer than the source signals. Consequently, prior work introduced the Half-VAE framework, which eliminates the encoder, thereby removing the explicit encoding and inverse mapping processes. This enables ICA or disentanglement tasks under underdetermined conditions to be optimized implicitly, bypassing the need for an explicit inverse mapping \(f^{-1}\). However, previous studies only demonstrated that Half-VAE performs comparably to traditional VAEs in determined conditions for ICA or disentanglement tasks, without directly addressing underdetermined scenarios.
    
    Both Half-VAE and traditional VAEs require carefully designed prior distributions for latent variables, with differences between priors encouraged to achieve independence, factorization, and structured representations. In this study, the Half Adversarial VAE (Half-AVAE) is proposed, incorporating adversarial networks and External Enhancement (EE) terms to enhance the independence of latent variable dimensions. Under underdetermined conditions, Half-AVAE infers source signals or independent components that closely align with the ground truth. This is a significant finding, as it validates the theoretical correctness of the Half-VAE framework when augmented with mechanisms to promote latent dimension independence, enabling the solution of underdetermined ICA or disentanglement tasks. This also demonstrates that the approximate posterior of latent variables can be directly optimized through the objective function, as shown in Equation (13). Furthermore, it underscores that the VAE framework is primarily driven by variational inference principles rather than a rigid encoder-decoder structure. Consequently, the encoder and decoder can be flexibly designed or omitted in algorithms combining variational inference and deep learning, provided the fundamental principles of variational inference are upheld.
    	
    \bibliography{ref.bib}

\end{document}